\documentclass[10pt,twocolumn,letterpaper]{article}

\usepackage[pagenumbers]{cvpr} 

\usepackage{times}
\usepackage{epsfig}
\usepackage{graphicx}
\usepackage{wrapfig}
\usepackage{amsmath}
\usepackage{amssymb}
\usepackage{multirow}
\usepackage{color, colortbl}
\usepackage{bm}
\usepackage{float}
\usepackage{pifont}
\usepackage{booktabs}
\usepackage{tikz}
\usepackage[super]{nth}
\usepackage[usestackEOL]{stackengine}
\usepackage[dvipsnames]{xcolor}

\definecolor{ArrowGreen}{RGB}{45,210,45}

\definecolor{Gray}{gray}{0.9}
\newcolumntype{g}{>{\columncolor{Gray}}c}

\makeatletter
\def\thickhline{%
  \noalign{\ifnum0=`}\fi\hrule \@height \thickarrayrulewidth \futurelet
   \reserved@a\@xthickhline}
\def\@xthickhline{\ifx\reserved@a\thickhline
               \vskip\doublerulesep
               \vskip-\thickarrayrulewidth
             \fi
      \ifnum0=`{\fi}}
\makeatother

\newlength{\thickarrayrulewidth}
\newcommand{\vect}[1]{\mathbf{#1}}
\DeclareMathOperator*{\argmin}{arg\,min}

\newcommand{\OURS}{Prism\xspace}
\newcommand{\greencheck}{{\color{ForestGreen}\ding{51}}}
\newcommand{\redcross}{{\color{BrickRed}\ding{55}}}
\newcommand{\Ell}{\mathcal{L}}

\newcommand\blfootnote[1]{%
  \begingroup
  \renewcommand\thefootnote{}\footnote{#1}%
  \addtocounter{footnote}{-1}%
  \endgroup
}

\setstackgap{L}{0.4cm}

\definecolor{cvprblue}{rgb}{0.21,0.49,0.74}
\usepackage[pagebackref,breaklinks,colorlinks]{hyperref}
\hypersetup{
  citecolor=ForestGreen,
  linkcolor=BrickRed,
}

\title{Prism: Semi-Supervised Multi-View Stereo with Monocular Structure Priors}
\author{
\Longunderstack{Alex Rich \\ \tt\small{anrich@ucsb.edu}}
\hspace{1cm}
\Longunderstack{Noah Stier \\ \tt\small{noahstier@ucsb.edu}}
\hspace{1cm}
\Longunderstack{Pradeep Sen \\ \tt\small{psen@ucsb.edu}}
\hspace{1cm}
\Longunderstack{Tobias H\"{o}llerer \\ \tt\small{holl@cs.ucsb.edu}}
\vspace{0.2cm}
\\
University of California, Santa Barbara
}

\begin{document}
\maketitle
\begin{abstract}
The promise of unsupervised multi-view-stereo (MVS) is to leverage large unlabeled datasets, yet current methods underperform when training on difficult data, such as handheld smartphone videos of indoor scenes.
Meanwhile, high-quality synthetic datasets are available but MVS networks trained on these datasets fail to generalize to real-world examples.
To bridge this gap, we propose a semi-supervised learning framework that allows us to train on real and rendered images jointly, capturing structural priors from synthetic data while ensuring parity with the real-world domain.
Central to our framework is a novel set of losses that leverages powerful existing monocular relative-depth estimators trained on the synthetic dataset, transferring the rich structure of this relative depth to the MVS predictions on unlabeled data.
Inspired by perceptual image metrics, we compare the MVS and monocular predictions via a deep feature loss and a multi-scale statistical loss.
Our full framework, which we call \OURS, achieves large quantitative and qualitative improvements over current unsupervised and synthetic-supervised MVS networks.
This is a best-case-scenario result, opening the door to using both unlabeled smartphone videos and photorealistic synthetic datasets for training MVS networks.
\end{abstract}    
\begin{figure}[t]
\begin{center}
    \includegraphics[width=\linewidth]{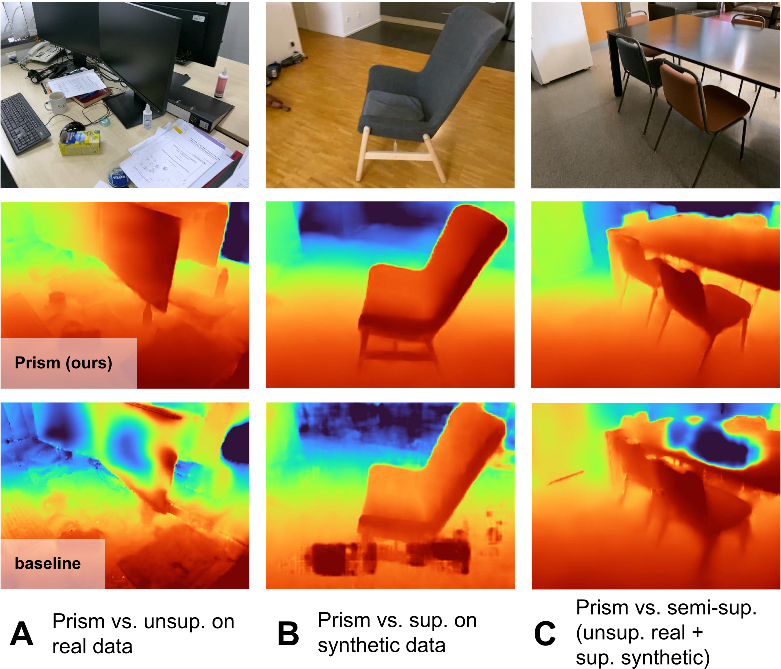}
\end{center}
\vspace{-0.4cm}
\begin{small}
\caption{Using structure priors from a monocular relative-depth network, \OURS effectively trains with a combination of real unlabeled smartphone video and synthetic data.
It outperforms all 3 baselines: unsupervised on smartphone data~(\textbf{A}), supervised on synthetic data~(\textbf{B}), and semi-supervised using both (\textbf{C}). Results shown are for the ScanNet++ dataset~\cite{dai2017scannet}. See Sec.~\ref{sec:exp} for details.}
\label{fig:teaser}
\end{small}
\vspace{-0.5cm}
\end{figure}
\blfootnote{\url{https://alexrich021.github.io/prism/}}

\section{Introduction} \label{sec:intro}
Multi-view stereo (MVS) is a classic and central problem in computer vision \cite{galliani2015gipuma, schoenberger2016mvs, tola2011mvs}, with applications from augmented reality to autonomous driving and robotics.
While fully-supervised deep-learning-based MVS has seen great advances \cite{cao2022mvsformer, ding2022trans, gu2020cas, liao2022wt, mi2022gbi, rich20213dvnet, yang2020cvp, yao2018mvs}, these methods rely heavily on accurate ground-truth 3D geometry collected using depth sensors.
This geometric information is time-consuming to collect, limiting the size of these datasets significantly compared to existing segmentation, classification, or text datasets, for example.

The promise of unsupervised multi-view stereo is to gain access to the same large amounts of training data available to other fields.
However, existing unsupervised MVS training methods \cite{chang2022rc, dai2019mvs2, khot2019unsup, huang2021m3vs, li2022ds, qiu2022pseudo, rich2024divloss, xiong2023cl, xu2021jdacs, xu2021umvs, yang2021sscvp, zhang2022elastic} generally only demonstrate training on the highly-constrained, laboratory-collected DTU dataset \cite{jensen2014large}.
We find that these methods cannot handle more difficult data such as handheld smartphone video of indoor scenes, often failing on reflective surfaces (Fig.~\ref{fig:teaser}{\color{BrickRed}A}) or complex geometry.
This significantly limits their real-world usefulness.
Concurrently, high-quality synthetic datasets have emerged~\cite{avetisyan2024scenescript, khanna2023hssd, roberts2021hypersim}, but their utility for training MVS networks is unclear. Networks trained fully-supervised with these datasets do not generalize to real examples, often predicting noisy surfaces and incorrect geometry (Fig.~\ref{fig:teaser}{\color{BrickRed}B}).
Using basic semi-supervision (i.e., jointly training unsupervised on smartphone data and supervised on synthetic data) does help performance, but these networks also predict incorrect geometry (Fig.~\ref{fig:teaser}{\color{BrickRed}C}).
Most notably, none of these baseline methods learn a reasonable \textit{structure} prior that matches the complexity of real data and therefore cannot handle, for instance, textureless and reflective surfaces ({\color{BrickRed}A}, {\color{BrickRed}C}) or thin structures ({\color{BrickRed}B}).

Meanwhile, recent work on diffusion-based monocular relative-depth predictors such as Marigold~\cite{ke2024marigold} and Lotus~\cite{he2024lotus} can train robust networks on small synthetic datasets.
These networks predict highly-structured but \textit{relative} depth, i.e., depth of arbitrary scale and shift from the ground truth.
Naturally, the question arises: can we transfer the structure prior these monocular networks learn on the synthetic data to MVS network predictions on real data?

To this end, we propose \textbf{\OURS}, a semi-supervised learning framework that leverages both unlabeled smartphone video and synthetic data effectively to train MVS networks with high-quality structure priors (Fig.~\ref{fig:teaser}, middle row).
Central to our framework is a novel set of losses that allows us to leverage powerful existing monocular relative-depth estimators trained on the synthetic dataset, transferring the rich structure of this relative depth to the MVS predictions on unlabeled data.
Our key observation with these losses is that we can apply ideas from RGB perceptual metrics in order to learn structure from relative depth.
Specifically, we apply two losses.
First, inspired by LPIPS~\cite{zhang2018lpips}, we use a pre-trained RGB feature extractor to extract deep features from the monocular and MVS predictions, which are then normalized and enforced to be close.
Second, inspired by multi-scale structural similarity~(SSIM)~\cite{wang2003msssim}, we encourage the statistics of the monocular and MVS predictions to be similar at multiple resolutions.
Interestingly, these concepts from perceptual image metrics apply readily to depth maps, allowing us to transfer structural priors from relative-depth networks and far outperforming pixel-wise $\ell_1$ and single-scale SSIM monocular losses.

In addition to these monocular losses, \OURS uses \mbox{unsupervised} losses on the smartphone data and supervised losses on synthetic examples.
We demonstrate joint training on ScanNet++ iPhone videos~\cite{yeshwanthliu2023scannetpp} and the Hypersim synthetic dataset~\cite{roberts2021hypersim}.
Our full framework achieves large quantitative and qualitative improvements in depth prediction over all baselines on the ScanNet++ test set~\cite{yeshwanthliu2023scannetpp}.
These results generalize to ARKitScenes~\cite{dehghan2021arkitscenes}, with \OURS again exceeding all baselines both quantitatively and qualitatively on all metrics.
Our contributions are as follows:
\begin{enumerate}
    \item We propose a deep feature loss and a multi-scale statistical loss for learning structure priors from relative depth and demonstrate their superior performance compared with existing monocular losses.
    \item We design \OURS, a semi-supervised learning framework which uses our monocular structure priors to train MVS networks on unlabeled real and labeled synthetic data.
    \item We demonstrate that \OURS outperforms all baseline methods: unsupervised on real data, supervised on synthetic data, and semi-supervised on both.
\end{enumerate}
\OURS opens the door to using both real-world videos and photorealistic synthetic datasets jointly for training MVS networks, taking an essential step towards more extensive or even unbounded training data for 3D reconstruction.
\section{Related Work}
\noindent{\textbf{Monocular priors:}}
Exploiting priors from monocular networks is common in many fields.
In sparse-view novel-view synthesis and neural-implicit 3D reconstruction, monocular cues are used to regularize ambiguous geometry. 
Generally, these cues are applied as pixel-wise constraints, using $\ell_1$ or $\ell_2$ losses~\cite{chung2024depthreg, hu2023consistentnerf, wang2023sparsenerf, yu2022monosdf} sometimes modulated by uncertainty estimations~\cite{roessle2022depthpriorsnerf, xiao2024debsdf}, multi-view checks~\cite{wang2022neuris}, or both~\cite{chen2024vcr}.
In monocular depth training, some methods train a teacher network, either with more parameters or on a separate dataset, and transfer the knowledge to a student network via pseudo-labels~\cite{wu2022practicalindoor, yang2024depthanythingv1, yang2024depthanythingv2}.
Pixel-wise supervision is most common here too~\cite{yang2024depthanythingv1, yang2024depthanythingv2}, though DistDepth~\cite{wu2022practicalindoor} proposes an SSIM loss, taking a step away from pixel-wise supervision.
In contrast to all of these methods, we transfer larger, patch-level \textit{structure} from monocular depth using both multi-scale statistics and deep features. We show this is much more effective than pixel-wise or single-scale SSIM approaches, making our work highly relevant.
Furthermore, to the best of our knowledge, monocular priors have not been applied to MVS depth prediction.

\noindent\textbf{Fully-supervised MVS:}
While supervised MVS depth prediction is a popular area of research~\cite{cao2022mvsformer, ding2022trans, duzceker2020dvmvs, hou2019gpmvs, im2019dps, liao2022wt, luo2019pmvs, rich20213dvnet, sayed2022simplerecon, xi2020ray, yao2018mvs, yi2020pyramid}, these methods remain constrained by their reliance on ground-truth 3D geometry.

\noindent\textbf{Unsupervised MVS:}
The field has largely settled on a combination of photometric, depth-smoothness, and augmentation-consistency losses~\cite{chang2022rc, dai2019mvs2, huang2021m3vs, khot2019unsup, li2022ds, rich2024divloss, xiong2023cl, xu2021jdacs}, sometimes with a 2-step, pseudo-labeling approach~\cite{ding2022kdmvs, qiu2022pseudo, xu2021umvs, yang2021sscvp}.
We demonstrate that these methods alone do not work on smartphone video, and propose a semi-supervised method that does work on this data.
Our framework allows these techniques to be widely applicable.

\noindent\textbf{Semi-supervised MVS:}
There is extremely limited work here. 
To the best of our knowledge, it all explores the single-dataset setting, either assuming sparse ground-truth points are provided~\cite{kim2021ssmvs, zhan2022ssmvs} or only a subset of images have ground-truth depth~\cite{xu2023ssmvs}.
We are the first work exploring the important area of multi-dataset semi-supervised MVS.

\noindent\textbf{Monocular depth prediction: }
While many works have focused on scaling the training set~\cite{bochkovskii2024depthpro, eftekhar2021omnidata, ranftl2022midas, li2018md, yang2024depthanythingv1, yang2024depthanythingv2}, Marigold~\cite{ke2024marigold} has recently proposed fine-tuning Stable Diffusion v2~\cite{rombach2022stable} for depth prediction on a small synthetic dataset. Lotus~\cite{he2024lotus} improves the inference time and training. We use these methods to capture structural priors from minimal synthetic data.
We show that we can transfer this prior to MVS predictions via carefully-constructed losses.
\begin{figure*}[t]
\begin{center}
    \includegraphics[width=\linewidth]{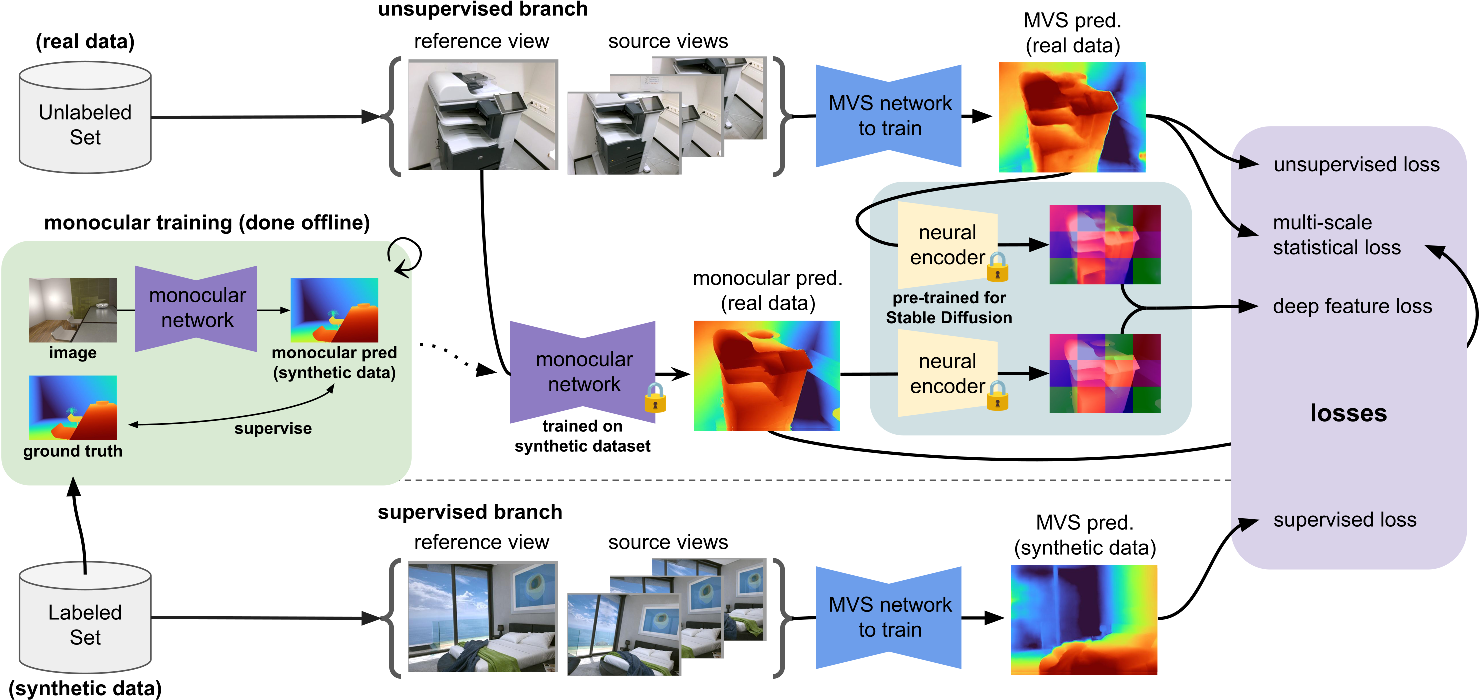}
\end{center}
\vspace{-0.4cm}
\begin{small}
\caption{\textbf{Overview of \OURS framework for semi-supervised MVS.} We leverage both real unlabeled smartphone data and labeled synthetic data to train MVS networks. Our central idea is to (1) train an existing monocular relative-depth prediction network on the synthetic set in order to capture high-quality structure priors and (2) teach the MVS network to use these structural priors on the unlabeled set via losses inspired by perceptual image metrics. In addition to these monocular losses, we also utilize unsupervised losses on the unlabeled real examples and supervised losses on the synthetic examples.}
\label{fig:main}
\end{small}
\vspace{-0.3cm}
\end{figure*}

\section{Method}
In this section we describe our novel \OURS framework for semi-supervised training of MVS networks (Fig.~\ref{fig:main}).
Our method takes as input an unlabeled and labeled set of images.
Each is assumed to have known camera parameters, and the labeled set is assumed to have ground-truth depth information. 
In our experiments, we use smartphone images and synthetic data for these sets.
Prior to training the MVS network, we train a monocular depth prediction network on the labeled set.
At each training iteration, reference images $\vect{I}^U$ and $\vect{I}^L$ are sampled from the unlabeled and labeled sets respectively, along with a set of source images and camera parameters and, for the labeled set, the ground-truth depth $\vect{D}^L_{gt}$. 
The MVS network to be trained makes predictions $\vect{D}^U$ and $\vect{D}^L$ on the unlabeled and labeled samples respectively. 
The loss is then computed as:
\begin{equation} \label{eq:full}
    \Ell_\mathrm{total} = \lambda_1 \Ell_\mathrm{mono} + \lambda_2 \Ell_\mathrm{unsup} + \lambda_3 \Ell_\mathrm{sup},
\end{equation}
where $\Ell_\mathrm{mono}$ uses the monocular network to compute a loss on $\vect{D}^U$, $\Ell_\mathrm{unsup}$ uses the standard photometric, depth-smoothness, and augmentation-consistency supervision from DIV loss~\cite{rich2024divloss} to compute a loss on $\vect{D}^U$, and $\Ell_\mathrm{sup}$ computes a supervised loss comparing $\vect{D}^L$ and $\vect{D}^L_{gt}$.
$\lambda_1, \lambda_2$, and $\lambda_3$ weight the contributions of each loss.
Other than differentiability, we make no assumptions on the MVS network to be trained.

\subsection{Monocular Structure Priors} \label{sec:mono}

As noted in Sec.~\ref{sec:intro} and Fig.~\ref{fig:teaser}, neither $\Ell_\mathrm{unsup}$ nor $\Ell_\mathrm{sup}$ help the MVS network learn reasonable structural priors for real examples.
To capture these priors from the labeled synthetic data, we train a monocular depth predictor.
We then teach the MVS network to use these structural priors on the unlabeled set via losses inspired by perceptual image metrics, specifically a deep feature loss inspired by LPIPS~\cite{zhang2018lpips} and a statistical loss inspired by multi-scale SSIM~\cite{wang2003msssim}.

\noindent{\textbf{Monocular network, prediction, and normalization:}}
For our monocular network, we take advantage of the methods which fine-tune diffusion models for relative-depth prediction on a small synthetic dataset~\cite{he2024lotus, ke2024marigold}.
Specifically, we train Marigold~\cite{ke2024marigold} on the labeled synthetic set prior to MVS training.
Then, given the reference image $\vect{I}^U$ on the unlabeled set, we make a relative-depth prediction $\vect{D}^U_*$ using this monocular network.
For later use with the deep feature loss, we normalize $\vect{D}^U_*$ to be in the range $[0, 1]$:
\begin{equation}
    \bar{\vect{D}}^U_* = \frac{\vect{D}^U_* - q_{2}}{q_{98} - q_{2}},
\end{equation}
where $q_{a}$ is the $a^{th}$ quantile of $\vect{D}^U_*$.
$\bar{\vect{D}}^U_*$ is affine-invariant, i.e., it is predicted up to a scale and shift ambiguity $s, t$ from the ground truth.
Following Ranftl \etal~\cite{ranftl2022midas}, we compute $s, t$ that align $\bar{\vect{D}}^U_*$ with the MVS prediction $\vect{D}^U$ as
\begin{equation} \label{eq:align}
    (s, t) = \argmin_{s, t} \sum_{\vect{p}} (s\bar{\vect{D}}^U_*(\vect{p}) + t - \vect{D}^U(\vect{p}))^2,
\end{equation}
which has an analytic solution.

\noindent{\textbf{Deep feature loss:}}
We want to transfer patch-level structure from the monocular depth prediction.
Inspired by LPIPS~\cite{zhang2018lpips}, we find deep features from pre-trained feature extractors are an extremely effective method of doing this.
First, these extractors naturally have a large receptive field.
Second, they are influenced more by higher-level structures than by pixel-level variation.

Specifically, we compare deep embeddings of $\vect{D}^U$ and $\bar{\vect{D}}^U_*$ using a pre-trained feature extractor.
This feature extractor expects 3-channel images with values in the range $[0, 1]$, so we first align $\vect{D}^U$ with $\bar{\vect{D}}^U_*$ as \mbox{$\bar{\vect{D}}^U = \frac{1}{s} (\vect{D}^U - t)$}, putting it approximately in the correct range.
We then duplicate both $H \times W$ depth maps $\bar{\vect{D}}^U$ and $\bar{\vect{D}}^U_*$ 3 times to form a $H \times W \times 3$ depth ``image." 
Using the pre-trained feature extractor, we compute $H' \times W' \times C$ deep embeddings $\vect{F}$ and $\vect{F}_*$ of these depth images.
Finally, we normalize in the channel dimension, denoting the normalized embeddings as $\bar{\vect{F}}$ and $\bar{\vect{F}}_*$, and take the mean $\ell_2$ distance between normalized embeddings as our feature loss $\ell_\mathrm{feat}$:
\begin{equation} \label{eq:feat}
    \ell_\mathrm{feat} = \frac{1}{H' W'} \sum_{\vect{p}} \| \bar{\vect{F}}(\vect{p}) - \bar{\vect{F}}_*(\vect{p}) \|.
\end{equation}
While we could compare multiple feature scales like LPIPS we find that for depth maps, unlike RGB images, comparing only the deepest feature embedding gives the best results.
We tested a variety of pre-trained feature extractors, and find that the encoder from Stable Diffusion v2~\cite{rombach2022stable} gives the largest performance boost. 
During initial development, we also tested passing our depth images directly to LPIPS, and found it to cause artifacts (see Sec.~\ref{sec:exp_abl}).

\noindent{\textbf{Statistical loss:}}
In addition to comparing deep features, we also find that comparing patch-wise statistics of $\vect{D}^U$ and $\vect{D}^U_*$ helps transfer the monocular structure prior.
There are two obvious options: SSIM~\cite{wang2002ssim, wang2004ssim} and MS-SSIM~\cite{wang2003msssim}.
In both cases, the loss can be taken as the negative similarity of aligned depth maps:
\begin{equation} \label{eq:ssim}
    \ell_\mathrm{ssim} = 1 - f_\mathrm{sim}(\vect{D}^U, s\bar{\vect{D}}_*^U + t),
\end{equation}
where $f_\mathrm{sim}$ denotes SSIM or MS-SSIM.
Single-scale SSIM does improve results slightly, but is limited to a single receptive field size.
MS-SSIM is computed as the product of patch-wise statistics for several patch scales.
This has the added benefit of multiple receptive fields increasing in size; however, we find the multiplication operation for combining the scale-wise statistics can cause instability in the loss term.
If one term is small, that term dominates and the other terms are ignored.
We find a summation operation to be more stable.
We therefore define a summation-based MS-SSIM as the normalized sum of single-scale SSIM:
\begin{equation} \label{eq:pyr}
    f_\mathrm{sim}(\vect{x}, \vect{y}) = \frac{1}{L} \sum_{l=1}^L 
    \mathrm{SSIM}(\downarrow_l(\vect{x}), \downarrow_l(\vect{y})).
\end{equation}
where $\downarrow_l(*)$ applies a low-pass filter and then downsamples to scale $l$.
Our loss $\ell_\mathrm{ssim}$ is then computed as in Eq.~\ref{eq:ssim}, the negative similarity.
Eq.~\ref{eq:pyr} simply computes SSIM on the image pyramid so, to distinguish it from standard MS-SSIM, we refer to it as ``pyramid SSIM" (P-SSIM).

\noindent{\textbf{Full monocular loss term:}}
Our final monocular loss is
\begin{equation}
    \Ell_\mathrm{mono} = \ell_\mathrm{feat} + \alpha \ell_\mathrm{ssim},
\end{equation}
where $\ell_\mathrm{feat}$ and $\ell_\mathrm{ssim}$ are given in Eqs.~\ref{eq:feat} and~\ref{eq:ssim} and $\alpha$ weights the terms. We set $\alpha = 1.0$ and use $L=4$ levels.

\subsection{Additional Losses} \label{sec:additional_losses}
In addition to $\Ell_\mathrm{mono}$, we also compute an unsupervised loss $\Ell_\mathrm{unsup}$ on $\vect{D}^U$ and supervised loss $\Ell_\mathrm{sup}$ on $\vect{D}^L$.

\noindent{\textbf{Unsupervised loss:}}
We use the standard photometric, depth-smoothness, and augmentation-consistency losses from the literature, opting for the exact DIV loss formulation recently proposed~\cite{rich2024divloss}.
The unsupervised loss is
\begin{equation}
    \Ell_\mathrm{unsup} = \alpha_1 \Ell_\mathrm{photo} + \alpha_2 \Ell_\mathrm{ssim} + \alpha_3 \Ell_\mathrm{sm} + \alpha_4 \Ell_\mathrm{aug}.
\end{equation}
We omit the details for space, and refer the reader to Rich~\etal~\cite{rich2024divloss}.
We use the same hyperparameters, multiplying $\alpha_3$ and $\alpha_4$ by a factor of 100 to account for dataset scale differences between DTU and ScanNet++.
Our final hyperparameters are $\alpha_1 = 12.0$, $\alpha_2 = 6.0$, $\alpha_3 = 18.0$, and $\alpha_4 = 1.0$.

\begin{figure*}[t]
\begin{center}
    \includegraphics[width=\linewidth]{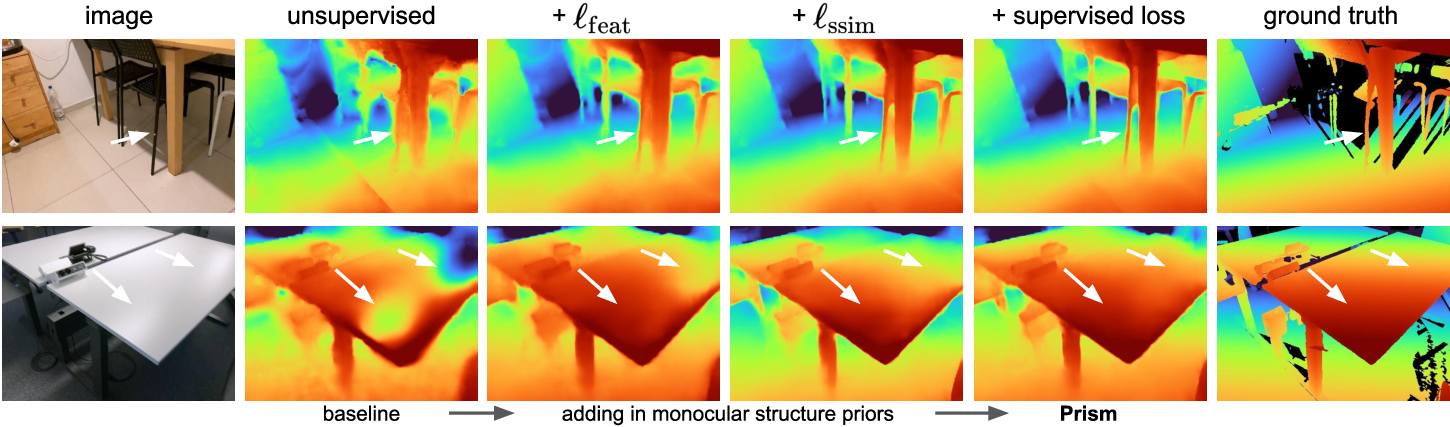}
\end{center}
\vspace{-0.4cm}
\begin{small}
\caption{\textbf{Visual Ablation Study.}
Each component of \OURS interacts constructively, helping bring out fine detail (top row) and global structure (bottom row).
The monocular structure prior significantly helps in the case of textureless/reflective surfaces. 
The supervised loss helps add a final smoothness to flat surfaces and sharpness to object boundaries.
See Sec.~\ref{sec:exp_abl} for details and Table~\ref{tab:ablation} for quantitative results.
}
\label{fig:abl}
\end{small}
\vspace{-0.5cm}
\end{figure*}

\noindent{\textbf{Supervised loss:}}
Inspired by Sayed \etal~\cite{sayed2022simplerecon}, we use a regression, multi-scale gradient, and normal loss:
\begin{equation}
    \Ell_\mathrm{sup} = \ell_\mathrm{regr} + \ell_\mathrm{grad} + \ell_\mathrm{normals}.
\end{equation}
The regression loss $\ell_\mathrm{regr}$ computes log $\ell_1$ error:
\begin{equation}
    \ell_\mathrm{regr} = \frac{1}{HW} \sum_\vect{p} |\log\vect{D}^L(\vect{p}) - \log\vect{D}^L_{gt}(\vect{p}) |.
\end{equation}
The multi-scale gradient loss $\ell_\mathrm{grad}$ is
\begin{equation}
    \ell_\mathrm{grad} = \sum_{l=1}^4 \sum_\vect{p} |\nabla\downarrow_l(\vect{D}^L(\vect{p})) - \nabla\downarrow_l(\vect{D}^L_{gt}(\vect{p}))|,
\end{equation}
where $\nabla$ is the first-order spatial gradient, and $\downarrow_l(*)$ applies a low-pass filter followed by downsampling to scale $l$, as before.
For $\ell_\mathrm{normals}$ we first extract normals $\vect{N}$ and $\vect{N}_{gt}$ from $\vect{D}^L$ and $\vect{D}^L_{gt}$ using the known camera intrinsics, then compute the loss as
\begin{equation}
    \ell_\mathrm{normals} = \frac{1}{2HW} \sum_{\vect{p}} 1 - \vect{N}(\vect{p})^T \vect{N}_{gt}(\vect{p}),
\end{equation}
where the 2 in the denominator normalizes the loss magnitude to $[0, 1]$.
In our experiments, we use CasMVSNet-style cost-volume-based networks~\cite{gu2020cas}. We find that specifically for the indoor setting we train and test on, these losses are superior to the $\ell_1$ regression or probability-based classification losses commonly used for these networks.
For the lower-resolution depth maps produced by our network, we upsample to full resolution and supervise only with $\ell_\mathrm{regr}$.

\begin{figure*}[t]
\begin{center}
    \includegraphics[width=\linewidth]{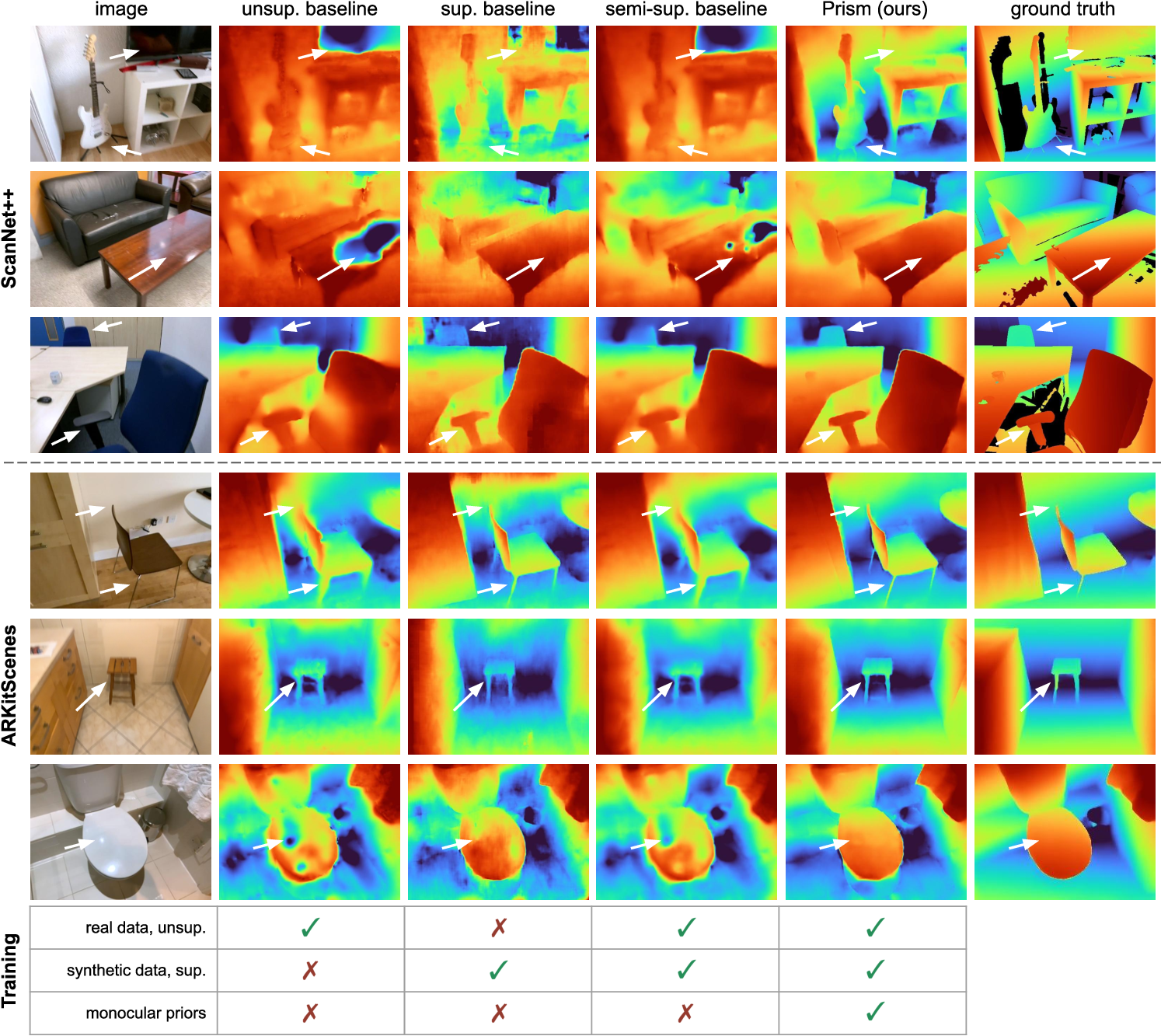}
\end{center}
\vspace{-0.4cm}
\begin{small}
\caption{\textbf{Qualitative Results.} \OURS outperforms all baselines, producing sharp and accurate depth maps with excellent global structure and fine-grained local detail. With the arrows, we indicate hard cases where \OURS performs well: textureless and reflective surfaces (rows 1, 2, 6) and thin structures (rows 1, 3, 4, 5). In our ablation study, we find the monocular losses largely help with textureless and reflective surfaces, while all components interact constructively to improve thin structures.}
\label{fig:qual}
\end{small}
\vspace{-0.2cm}
\end{figure*}
\subsection{Final Details}
We set the monocular, unsupervised, and supervised weights as $\lambda_1~=~10, \lambda_2~=~1$, and $\lambda_3 = 10$.
The unsupervised loss is approximately 10x the magnitudes of the others, so we chose these hyperparameters to balance the three.
It is likely better hyperparameters can be found using a validation set; however, we find these to be quite effective.
Finally, we only activate the monocular loss after the first epoch of training, allowing for initial convergence so the scale and shift estimation (Eq.~\ref{eq:align}) is reasonable.

\begin{table*}[t]
\setlength\tabcolsep{6pt}
\begin{small}
\begin{center}
\begin{tabular}{r c c c c c|g c c c}
\thickhline
    & \multirow{2}{*}{RC \cite{chang2022rc}} & \multirow{2}{*}{CL \cite{xiong2023cl}} & \multirow{2}{*}{DIV~\cite{rich2024divloss}} & \multirow{2}{*}{Cas~\cite{gu2020cas}} & DIV~\cite{rich2024divloss} & \textbf{\OURS} & \multicolumn{3}{c}{\underline{\% diff.~vs.~baselines}} \\
    & & & & & + sup. & (ours) & vs.~unsup. & vs.~sup. & vs.~semi-sup. \\
\thickhline
\multicolumn{1}{l}{\sc{Training}} & & & {\scriptsize(unsup.~base)} & {\scriptsize(sup.~base)} & {\scriptsize(semi-sup.~base)} &\\
    real data, unsup. & \greencheck & \greencheck & \greencheck & \redcross & \greencheck & \greencheck \\
    synthetic, sup. & \redcross & \redcross & \redcross & \greencheck & \greencheck & \greencheck \\
    monocular priors & \redcross & \redcross & \redcross & \redcross & \redcross & \greencheck \\
\thickhline
\multicolumn{1}{l}{\sc{ScanNet++}} & & & & & & & & & \\
Abs-rel $\downarrow$ & 0.124 & 0.126 & 0.123 & 0.118 & 0.100 & \textbf{0.090} & {\color{ForestGreen}-26.8\%} & {\color{ForestGreen}-23.7\%} & {\color{ForestGreen}-10.0\%}\\
Abs-diff $\downarrow$ & 0.207 & 0.222 & 0.200 & 0.215 & 0.179 & \textbf{0.158} & {\color{ForestGreen}-21.0\%} & {\color{ForestGreen}-26.5\%} & {\color{ForestGreen}-11.7\%}\\
Abs-inv $\downarrow$ & 0.090 & 0.091 & 0.088 & 0.133 & 0.077 & \textbf{0.068} & {\color{ForestGreen}-22.7\%} & {\color{ForestGreen}-48.9\%} & {\color{ForestGreen}-11.7\%}\\
Sq-rel $\downarrow$ & 0.082 & 0.082 & 0.086 & 0.086 & 0.066 & \textbf{0.052} & {\color{ForestGreen}-39.5\%} & {\color{ForestGreen}-39.5\%} & {\color{ForestGreen}-21.2\%}\\
RMSE $\downarrow$ & 0.323 & 0.335 & 0.316 & 0.335 & 0.288 & \textbf{0.250} & {\color{ForestGreen}-20.9\%} & {\color{ForestGreen}-25.4\%} & {\color{ForestGreen}-13.2\%}\\
$\delta < 1.25$ $\uparrow$ & 0.840 & 0.829 & 0.848 & 0.831 & 0.873 & \textbf{0.895} & {\color{ForestGreen}5.5\%} & {\color{ForestGreen}7.7\%} & {\color{ForestGreen}2.5\%}\\
\thickhline
\multicolumn{1}{l}{\sc{ARKitScenes}} & & & & & & \\
    Abs-rel $\downarrow$ & 0.193 & 0.182 & 0.182 & 0.129 & 0.127 & \textbf{0.115} & {\color{ForestGreen}-36.8\%} & {\color{ForestGreen}-10.9\%} & {\color{ForestGreen}-9.4\%}\\
    Abs-diff $\downarrow$ & 0.215 & 0.212 & 0.203 & 0.183 & 0.165 & \textbf{0.148} & {\color{ForestGreen}-27.1\%} & {\color{ForestGreen}-19.1\%} & {\color{ForestGreen}-10.3\%}\\
    Abs-inv $\downarrow$ & 0.144 & 0.137 & 0.138 & 0.169 & 0.108 & \textbf{0.100} & {\color{ForestGreen}-27.5\%} & {\color{ForestGreen}-40.8\%} & {\color{ForestGreen}-7.4\%}\\
    Sq-rel $\downarrow$ & 0.154 & 0.139 & 0.163 & 0.077 & 0.096 & \textbf{0.069} & {\color{ForestGreen}-57.7\%} & {\color{ForestGreen}-10.4\%} & {\color{ForestGreen}-28.1\%}\\
    RMSE $\downarrow$ & 0.322 & 0.313 & 0.308 & 0.275 & 0.258 & \textbf{0.224} & {\color{ForestGreen}-27.3\%} & {\color{ForestGreen}-18.5\%} & {\color{ForestGreen}-13.2\%}\\
    $\delta < 1.25$ $\uparrow$ & 0.812 & 0.815 & 0.824 & 0.829 & 0.873 & \textbf{0.890} & {\color{ForestGreen}8.0\%} & {\color{ForestGreen}7.4\%} & {\color{ForestGreen}1.9\%}\\
\thickhline
\end{tabular}
\end{center}
\vspace{-0.4cm}
\caption{\textbf{Main Comparisons.} Depth prediction results on ScanNet++~\cite{yeshwanthliu2023scannetpp} and ARKitScenes~\cite{dehghan2021arkitscenes}. We compare with several unsupervised methods (RC, CL, DIV), as well as a fully-supervised baseline trained on the synthetic Hypersim dataset~\cite{roberts2021hypersim}, and a semi-supervised baseline without our novel monocular structure priors. \OURS outperforms all baselines on all metrics on both datasets, demonstrating the efficacy of our monocular structure priors. In particular, we find metrics are improved by over {\color{ForestGreen}\textbf{10\%}} in nearly all cases vs.~our semi-supervised baseline. We outperform the published unsupervised and synthetic-supervised baselines by over {\color{ForestGreen}\textbf{20}} to {\color{ForestGreen}\textbf{30\%}} on many metrics.}
\label{tab:all}
\end{small}
\end{table*}

\section{Experiments} \label{sec:exp}
\subsection{Implementation Details}
For our MVS network, we follow DIV-MVS~\cite{rich2024divloss}: CasMVSNet~\cite{gu2020cas} with group-wise correlation~\cite{xu2019cider} and the pixel-wise weight map for cost-volume aggregation.
We implement Prism in PyTorch~\cite{pytorch}. 
Additionally, we make heavy use of Open3D~\cite{Zhou2018o3d} for visualization.

\noindent{\textbf{Training data:}} For our unlabeled smartphone dataset, we use the ScanNet++ iPhone video training split~\cite{yeshwanthliu2023scannetpp} which consists of handheld capture of 230 indoor scenes.
We sample every 10 frames from the iPhone videos.
The data comes with corresponding registered and meshed laser scans, so ground-truth depth can be rendered for each frame; however, there is no in-line depth sensor and the rendered depth often contains artifacts.
For our labeled synthetic dataset, we use the Hypersim training split~\cite{roberts2021hypersim}, which consists of high-quality, photo-realistic renderings of 365 artist-created indoor spaces with corresponding depth ground truth.
We resize all images to $384 \times 512$.

\noindent{\textbf{Baselines:}}
We compare against 3 main baselines.
The first is our base unsupervised method, DIV-MVS~\cite{rich2024divloss}.
The second is our base supervised method, i.e., our CasMVSNet-style network trained using $\Ell_\mathrm{sup}$ on synthetic data.
The third is a base semi-supervised method trained both unsupervised on real data and supervised on synthetic data but without monocular structure priors (i.e., using only $\Ell_\mathrm{unsup}$ and $\Ell_\mathrm{sup}$ but not $\Ell_\mathrm{mono}$).
Note that for all of these baselines, we keep the MVS network constant and change \textit{only} the training.
Furthermore, to the best of our knowledge, our semi-supervised baseline represents a novel combination of joint supervised and unsupervised MVS training. Due to its good performance, we hope that it may be useful for future work.
In addition to these main baselines, we also compare against two other unsupervised methods from the literature, RC-MVSNet~\cite{chang2022rc} and CL-MVSNet~\cite{xiong2023cl}.

\noindent{\textbf{Training parameters:}} All pipelines are trained from scratch for 160k steps. We use the Adam optimizer~\cite{kingma2017adam}, with an initial learning rate of $10^{-4}$ and a weight decay of $10^{-4}$. The learning rate is halved after 100k, 120k, and 140k steps.
We use a batch size of 4, which requires 4 NVIDIA RTX 3090 GPUs.
For all unsupervised baselines, we use the hyperparameters detailed in Sec.~\ref{sec:additional_losses}. 
Following existing work~\cite{chang2022rc, rich2024divloss, xu2021jdacs}, we double the augmentation-consistency weight every 20k steps from step 10k until 90k.

\subsection{Evaluation Details}
\noindent{\textbf{Testing data:}} For our main comparisons, we use the \mbox{ScanNet++} iPhone semantic test set~\cite{yeshwanthliu2023scannetpp}, which consists of 50 complex and challenging indoor scenes.
To test the generalization ability of \OURS, we additionally test on the \mbox{ARKitScenes} 3D object detection validation set~\cite{dehghan2021arkitscenes} with \textit{no finetuning on additional data}.
This dataset consists of iPhone images with corresponding depth maps from ARKit.
We randomly select 50 scenes from the subset of the validation set where ``up" in the image is aligned with gravity.

\noindent{\textbf{Protocols:}}
We use 5 images for depth prediction.
For both datasets, we resize the images to $384 \times 512$ and make a depth prediction at the same resolution.
We set the minimum and maximum depth planes to $25\mathrm{cm}$ and $5\mathrm{m}$ respectively for both datasets.
For quantitative comparison, we use the standard metrics from Murez~\etal~\cite{murez2020atlas}.
Abs-rel and Abs-diff error are considered the most important among these.

\subsection{Results}
\begin{figure*}[t]
\begin{center}
    \includegraphics[width=\linewidth]{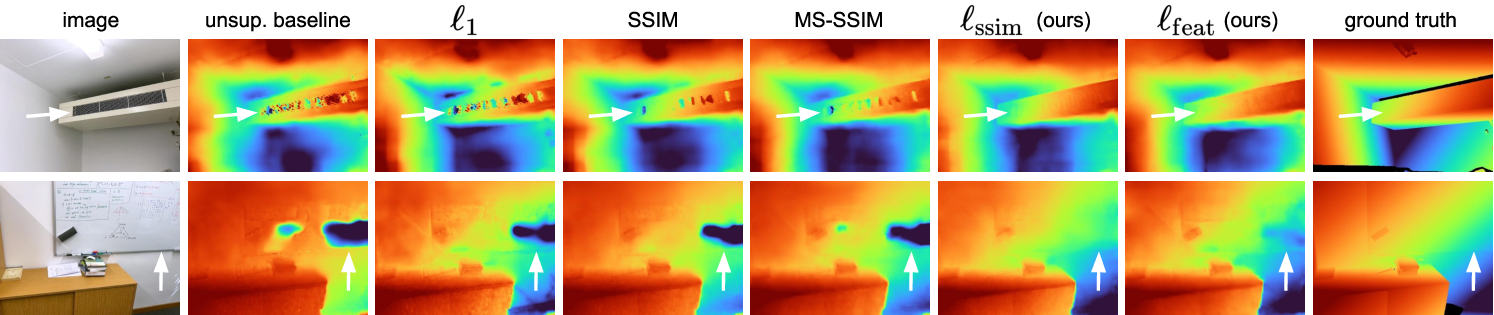}
\end{center}
\vspace{-0.3cm}
\begin{small}
\caption{\textbf{Visual Ablation Study.} We isolate the effect of various monocular losses. Only our multi-scale loss $\ell_\mathrm{ssim}$ and our deep feature loss $\ell_\mathrm{feat}$ can handle confusing geometry (top row) and textureless surfaces (bottom row). See Table~\ref{tab:ablation} for quantitative results.}
\label{fig:mono_abl}
\end{small}
\vspace{-0.5cm}
\end{figure*}
Overall, we see \OURS far outperforms competing baselines on all datasets tested (Sec.~\ref{sec:exp_comp}) and each component of \OURS interacts constructively to improve results (Sec.~\ref{sec:exp_abl}).

\subsection{Main Comparisons}~\label{sec:exp_comp}
See Table~\ref{tab:all} and Fig.~\ref{fig:qual} for quantitative and qualitative results on ScanNet++ and ARKitScenes.
\OURS outperforms all competing baselines on all metrics on both datasets~\mbox{(Table~\ref{tab:all})}. In particular, we improve metrics by over~\mbox{\color{ForestGreen}\textbf{10\%}} in nearly all cases compared to our semi-supervised baseline, confirming the effectiveness of our monocular structure priors.
We highlight that this is true not only for ScanNet++, whose training data we use for our unlabeled set, but also for \mbox{ARKitScenes}.
This indicates \OURS learns a general structure prior that transfers across datasets.
We also note that when comparing exclusively to published work, i.e., the unsupervised and synthetic-supervised baselines, we improve many metrics by over {\color{ForestGreen}\textbf{20}} to {\color{ForestGreen}\textbf{30\%}}.

Qualitatively, \OURS produces depth maps with coherent global structure and sharp local detail (Fig.~\ref{fig:qual}).
In particular, we find \OURS noticeably improves performance in two hard cases.
First, \OURS predicts accurate depth for textureless and reflective surfaces while most competing baselines fail.
Second, \OURS predicts thin structures with high precision while competing baselines tend to predict blurry, overly-smoothed object boundaries.
These results further highlight the effectiveness of our monocular structure priors.


\subsection{Ablation Study} \label{sec:exp_abl}
In Table~\ref{tab:ablation}, we analyze each component of \OURS, showing our monocular structure priors are critical to our performance boost and each component interacts constructively.
Note that rows 0, 1, and 2 of the table correspond to our unsupervised, supervised, and semi-supervised base methods.

\noindent{\textbf{Basic semi-supervision helps:}}
We first briefly note that basic semi-supervision (row 2) does provide a nice quantitative boost.
However, as detailed in Sec.~\ref{sec:exp_comp} and shown in Figs.~\ref{fig:teaser} and~\ref{fig:qual}, it fails to learn a structure prior, often failing on textureless/reflective surfaces and thin structures.
\begin{table*}[t]
\setlength\tabcolsep{4pt}
\begin{small}
\begin{center}
\begin{tabular}{c|c c c|c c c c|c c c c}
\thickhline
& \multicolumn{3}{c|}{\underline{losses}} & \multicolumn{4}{c|}{ScanNet++} & \multicolumn{4}{c}{ARKitScenes} \\
& unsup. & sup. & mono. & Abs-rel $\downarrow$ & Abs-diff $\downarrow$ & Abs-inv $\downarrow$ & $\delta < 1.25$ $\uparrow$ & Abs-rel $\downarrow$ & Abs-diff $\downarrow$ & Abs-inv $\downarrow$ & $\delta < 1.25$ $\uparrow$ \\
\thickhline
    0 & \checkmark & & & 0.123 & 0.200 & 0.088 & 0.848 & 0.182 & 0.203 & 0.138 & 0.824 \\
    1 & & \checkmark & & 0.118 & 0.215 & 0.133 & 0.831 & 0.129 & 0.183 & 0.169 & 0.829 \\
    2 & \checkmark &\checkmark & & 0.100 & 0.179 & 0.077 & 0.873 & 0.127 & 0.165 & 0.108 & 0.873 \\
    3 & \checkmark & & L1 & 0.110 & 0.206 & 0.083 & 0.852 & 0.150 & 0.188 & 0.126 & 0.841 \\
    4 & \checkmark & & SSIM & 0.107 & 0.193 & 0.081 & 0.859 & 0.148 & 0.181 & 0.123 & 0.850 \\
    5 & \checkmark & & MS-SSIM & 0.100 & 0.181 & 0.075 & 0.875 & 0.136 & 0.171 & 0.114 & 0.867 \\
    6 & \checkmark & & $\ell_\mathrm{ssim}$ & 0.096 & 0.178 & 0.074 & 0.880 & 0.127 & 0.165 & 0.114 & 0.867 \\
    7 & \checkmark & & $\ell_\mathrm{feat}$ & 0.099 & 0.171 & 0.072 & 0.884 & 0.135 & 0.162 & 0.110 & 0.875 \\
    8 & \checkmark & & $\ell_\mathrm{feat} + \ell_\mathrm{ssim}$ & 0.093 & 0.165 & 0.070 & 0.889 & 0.123 & 0.154 & 0.108 & 0.879 \\
    9 & \checkmark & \checkmark & $\ell_\mathrm{feat} + \ell_\mathrm{ssim}$ & \textbf{0.090} & \textbf{0.158} & \textbf{0.068} & \textbf{0.895} & \textbf{0.115} & \textbf{0.148} & \textbf{0.100} & \textbf{0.890} \\
\hline
\end{tabular}
\end{center}
\vspace{-0.3cm}
\caption{\textbf{Ablation Study.} We show each component of \OURS interacts constructively and our monocular structure priors noticeably outperform existing options. See Sec.~\ref{sec:exp_abl} for details and Figs.~\ref{fig:abl} and~\ref{fig:mono_abl} for qualitative results.}
\label{tab:ablation}
\end{small}
\vspace{-0.1cm}
\end{table*}
\begin{figure}[t]
\begin{center}
    \includegraphics[width=\linewidth]{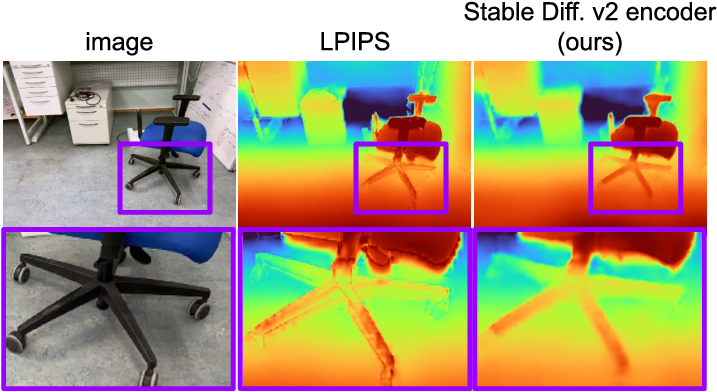}
\end{center}
\vspace{-0.3cm}
\begin{small}
\caption{\textbf{LPIPS causes artifacts.}
We initially tried LPIPS~\cite{zhang2018lpips}, and found that it causes spurious artifacts. Though these look like sharp details, they are actually hallucinations. Our solution alleviates this problem. See Sec~\ref{sec:exp_abl} for details and discussion.
}
\label{fig:artifacts}
\end{small}
\vspace{-0.2cm}
\end{figure}
\begin{table}[t]
\setlength\tabcolsep{5pt}
\begin{small}
\begin{center}
\begin{tabular}{r c c c c}
\thickhline
    & unsup. & Omni- & Marigold & Marigold \\ 
    & base & data & {\scriptsize(pre-trained)} & {\scriptsize(ours)} \\
\thickhline
\multicolumn{1}{l}{\sc{ScanNet++}} & & & \\
    Abs-rel $\downarrow$ & 0.123 & 0.093 & 0.095 & 0.093 \\
    Abs-diff $\downarrow$ & 0.200 & 0.163 & 0.165 & 0.165 \\
\thickhline
\multicolumn{1}{l}{\sc{ARKitScenes}} & & & \\
    Abs-rel $\downarrow$ & 0.182 & 0.121 & 0.126 & 0.123\\
    Abs-diff $\downarrow$ & 0.203 & 0.152 & 0.156 & 0.154 \\
\thickhline
\end{tabular}
\end{center}
\vspace{-0.4cm}
\caption{\textbf{Our method works with other monocular networks.} We test with two pre-trained networks. Results shown do not use the supervised loss. See Sec.~\ref{sec:exp_abl} for details.}
\label{tab:mono}
\end{small}
\vspace{-0.3cm}
\end{table}

\noindent{\textbf{Our monocular structure priors are critical:}} 
In rows 3 through 7 of Table~\ref{tab:ablation}, we show our monocular losses outperform existing methods.
The baseline method for these rows is unsupervised (row 0).
We test 3 alternative options: the commonly used pixel-wise $\ell_1$ loss (row 3), as well as SSIM (row 4) and MS-SSIM (row 5).
$\ell_1$ and SSIM provide limited benefit.
Only with MS-SSIM does performance begin to noticeably improve, indicating that reasonably-sized receptive fields help these losses.
Our $\ell_\mathrm{ssim}$ (row 6, Eq.~\ref{eq:ssim}) improves on the standard MS-SSIM, boosting performance in most cases while maintaining performance otherwise.
Our deep-feature loss (row 7, $\ell_\mathrm{feat}$, Eq.~\ref{eq:feat}) provides a similar boost as $\ell_\mathrm{ssim}$ and outperforms MS-SSIM on every metric.

In Fig.~\ref{fig:mono_abl}, we isolate the effect of each monocular loss.
Only our proposed losses perform well on confusing geometry (top row) or textureless/reflective surfaces (bottom row), further confirming our choices.

\noindent{\textbf{\OURS is greater than the sum of its parts:}}
In rows~8 and~9 of Table~\ref{tab:ablation} and Fig.~\ref{fig:abl}, we demonstrate that each component of \OURS interacts constructively.
In row 8, we show using $\ell_\mathrm{ssim}$ and $\ell_\mathrm{feat}$ together boosts performance on every metric.
In row 9, we show the supervised loss improves performance beyond both semi-sup.~(row~2) and unsup.~with monocular priors (row 8).
In Fig.~\ref{fig:abl}, we demonstrate that each component cumulatively improves performance on thin structures and textureless/reflective surfaces.

\noindent{\textbf{LPIPS causes artifacts:}}
While developing our deep feature loss, we tried using LPIPS~\cite{zhang2018lpips} with VGG~\cite{simonyan15vgg}.
As shown in Fig.~\ref{fig:artifacts}, it causes incorrect depth at object boundaries.
Though initially these look like sharp details, they are actually hallucinations.
We saw this effect across numerous examples.
Our solution alleviates this problem.

\noindent{\textbf{Our method works with other monocular networks:}}
In Table~\ref{tab:mono}, we test two pre-trained monocular networks: Omnidata~\cite{eftekhar2021omnidata}, and pre-trained Marigold~\cite{ke2024marigold} (the network architecture we use). To study only the monocular network, we remove $\Ell_\mathrm{sup}$. Our structural losses similarly improve results independent of network choice, indicating general applicability not tied to specific architectures.
Furthermore, we get a slight boost in performance by removing the out-of-domain VKITTI2 driving dataset~\cite{cabon2020vkitti2} that is in the original Marigold training set and training only on the indoor (i.e., in-domain) Hypersim dataset.

\section{Conclusions}
We have proposed \OURS, a novel semi-supervised learning framework that allows us to train MVS networks on real and rendered images jointly.
Central to our framework is a novel set of losses that leverages powerful existing monocular relative-depth estimators trained on the synthetic dataset, transferring the rich \textit{structure} of this relative depth to the MVS predictions on unlabeled data.
Inspired by perceptual image metrics, these losses consist of a deep feature loss and a multi-scale statistical loss.
We have demonstrated that both outperform existing monocular losses while also interacting constructively.
Our work bridges the gap between training on real-world RGB videos and photorealistic synthetic datasets, taking an essential step towards more extensive or even unbounded training data for 3D reconstruction.
{
    \small
    \bibliographystyle{ieeenat_fullname}
    \bibliography{main}
}


\end{document}


\maketitlesupplementary

\section{Overview}

In this supplementary material, we include additional details to complement the main text.
See Sec.~\ref{sec:addl_qual}, we include additional qualitative results.
In Sec.~\ref{sec:addl_exp}, we include additional experimental results showing \OURS works with the popular MVSFormer~\cite{cao2022mvsformer} architecture.
In Sec.~\ref{sec:metrics}, we include definitions for our depth-prediction metrics.

\section{Additional Qualitative Results} \label{sec:addl_qual}
\begin{figure*}[t]
\begin{center}
    \includegraphics[width=\linewidth]{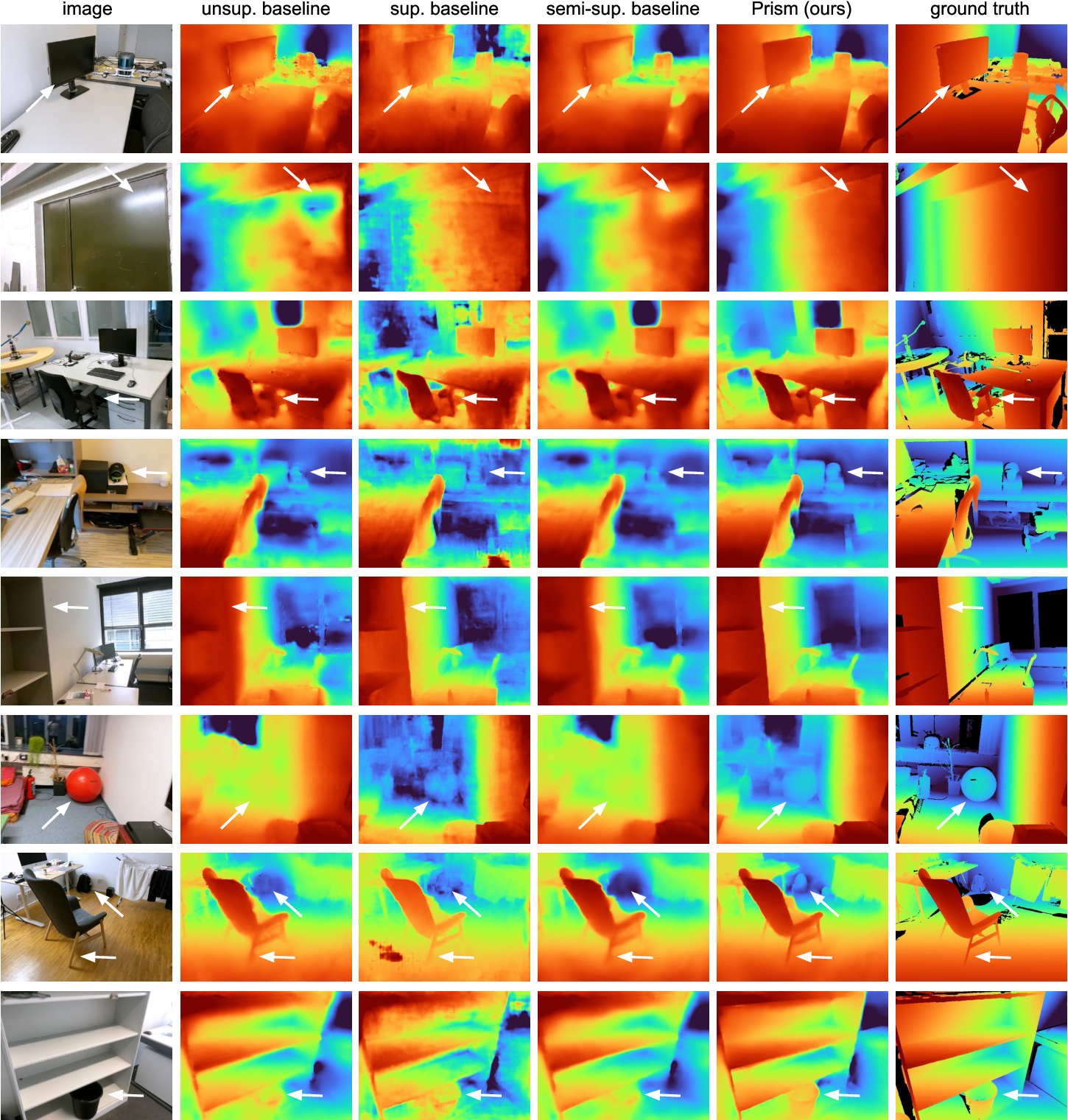}
\end{center}
\begin{small}
\caption{\textbf{Additional Qualitative Results (ScanNet++).} We include additional depth-prediction results to supplement the main text. \OURS again outperforms all baselines, producing sharp and accurate depth maps with excellent global structure and fine-grained local detail.
With the arrows, we indicate example hard cases where \OURS performs well: objects far in the background (rows 4, 6, 7), textureless and reflective surfaces (rows 1, 2, 8) and thin structures/object boundaries (rows 3, 5, 7). See Sec.~\ref{sec:addl_qual} for details. For details regarding the baseline methods, see Sec.~4.1 of the main text.}
\label{fig:addl_snpp}
\end{small}
\end{figure*}
\begin{figure*}[t]
\begin{center}
    \includegraphics[width=\linewidth]{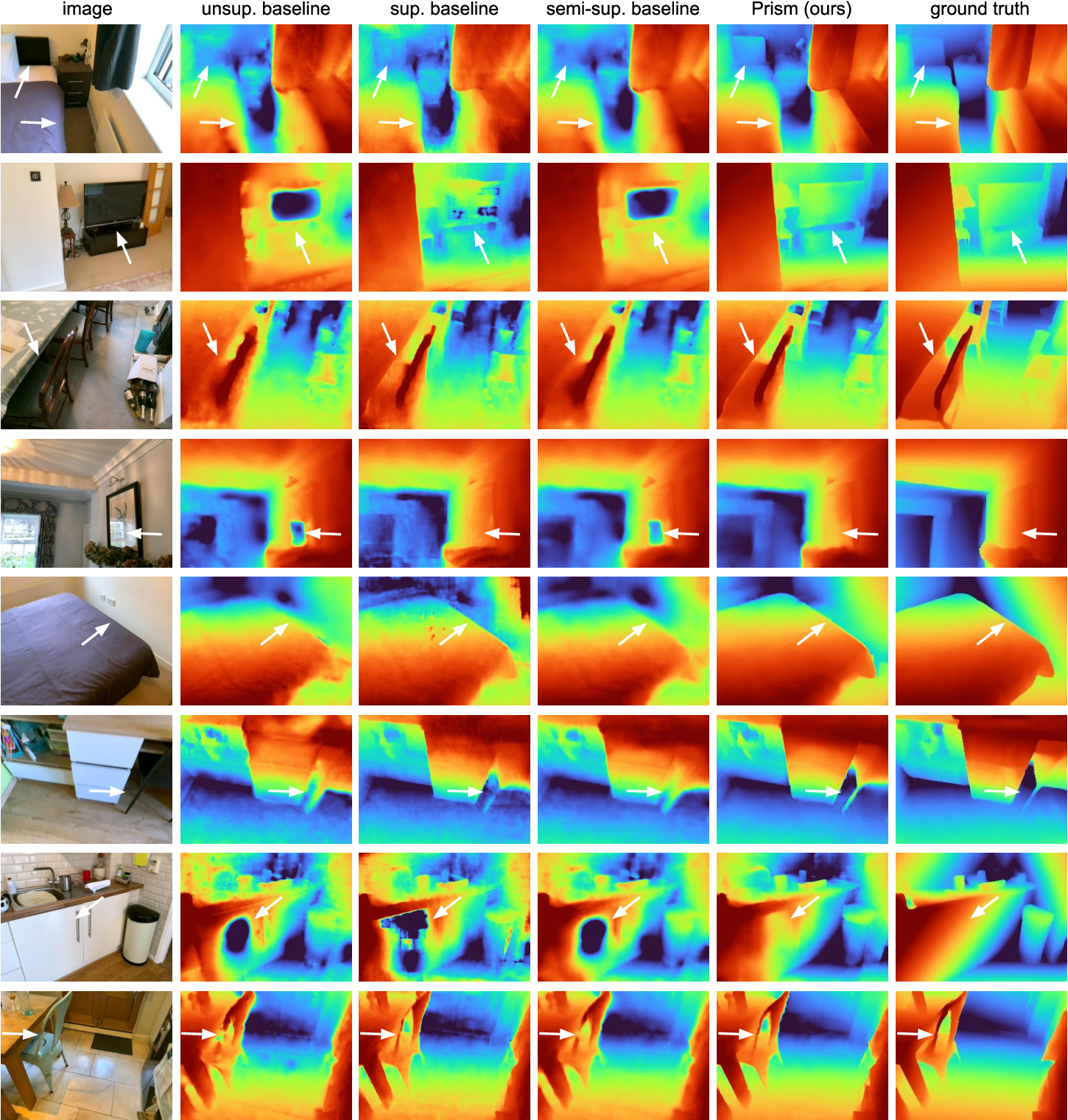}
\end{center}
\begin{small}
\caption{\textbf{Additional Qualitative Results (ARKitScenes).} We include additional depth-prediction results to supplement the main text. \OURS generalizes to the ARKitScenes dataset, outperforming all competing baselines.
As with ScanNet++ (Fig.~\ref{fig:addl_snpp}), we find \OURS performs well on many hard cases: objects far in the background (row 1), textureless/reflective surfaces (rows 2, 4, 7), and thin structures/object boundaries (rows 1, 3, 5, 6, 8). See Sec.~\ref{sec:addl_qual} for details. For details regarding the baseline methods, see Sec.~4.1 of the main text.}
\label{fig:addl_arks}
\end{small}
\end{figure*}
In Figs.~\ref{fig:addl_snpp} and \ref{fig:addl_arks}, we include additional qualitative depth-prediction results on ScanNet++ and ARKitScenes.
These results demonstrate, as in the main text, that \OURS noticeably boosts performance on two hard cases where most competing baselines fail: textureless/reflective surfaces and thin structures/object boundaries.
In addition, we also include examples of \OURS making sharp and accurate predictions for objects deep in the background of the scene.
In comparison, we find competing baselines perform poorly for these cases, often predicting blurry/noisy objects or sometimes not even predicting the object at all.
For a good example of this final case, see Fig.~\ref{fig:addl_snpp} row 7.

These additional results demonstrate that the qualitative improvements of \OURS extend to numerous examples across both evaluation datasets.
This further underscores the effectiveness of our monocular structure priors.
In all experiments, we found our structure priors to be critical for these improvements.

\section{\OURS works with MVSFormer} \label{sec:addl_exp}
In our main experiments, we used the CasMVSNet-style base network from DIV-MVS.
We chose this model for fair comparison with existing unsupervised MVS methods.
In Table~\ref{tab:mvsformer}, we include results instead using the \mbox{MVSFormer-P} model~\cite{cao2022mvsformer}.
This model leverages a pre-trained Dino~\cite{caron2021dino} network to augment the 2D feature extraction step, and is a highly popular MVS network.

We find that \OURS works seamlessly with the MVSFormer model.
As expected, every metric is improved with this more advanced model compared to \OURS with the CasMVSNet-style network.
This is a highly-relevant result, indicating general applicability of \OURS not tied to a specific MVS network architecture.
\begin{table}[h]
\setlength\tabcolsep{8pt}
\begin{small}
\begin{center}
\begin{tabular}{r c c}
\thickhline
    & \multirow{2}{*}{\textbf{\OURS} (ours)} & \textbf{\OURS} (ours) \\ 
    & & + MVSFormer \\
\thickhline
\multicolumn{1}{l}{\sc{ScanNet++}} & & \\
    Abs-rel $\downarrow$ & 0.090 & 0.083 \\
    Abs-diff $\downarrow$ & 0.158 & 0.151 \\
    Abs-inv $\downarrow$ & 0.068 & 0.063 \\
    Sq-rel $\downarrow$ & 0.052 & 0.042 \\
    RMSE $\downarrow$ & 0.250 & 0.237 \\
    $\delta < 1.25$ $\uparrow$ & 0.895 & 0.905 \\
\thickhline
\multicolumn{1}{l}{\sc{ARKitScenes}} & & \\
    Abs-rel $\downarrow$ & 0.115 & 0.104 \\
    Abs-diff $\downarrow$ & 0.148 & 0.144 \\
    Abs-inv $\downarrow$ & 0.100 & 0.094 \\
    Sq-rel $\downarrow$ & 0.069 & 0.063 \\
    RMSE $\downarrow$ & 0.224 & 0.218 \\
    $\delta < 1.25$ $\uparrow$ & 0.890 & 0.899 \\
\thickhline
\end{tabular}
\end{center}
\vspace{-0.2cm}
\caption{\textbf{\OURS works with MVSFormer.} We test \OURS with the popular MVSFormer-P network~\cite{cao2022mvsformer} in place of our CasMVSNet-style base network. We find that \OURS works seamlessly with MVSFormer. As expected with this more advanced model, every metric is improved.
This is an important result, indicating general applicability of \OURS not tied to a specific MVS network architecture.
See Sec.~\ref{sec:addl_exp} for details.}
\label{tab:mvsformer}
\end{small}
\end{table}

\section{Metric Definitions} \label{sec:metrics}
We use the standard depth-prediction metrics from Murez \etal~\cite{murez2020atlas}. See Table~\ref{tab:metrics} for the definitions.
Of these metrics, Abs-rel and Abs-diff are considered the most important.
\begin{table}[h]
\setlength\tabcolsep{10pt}
\begin{small}
\begin{center}
\begin{tabular}{r|c}
\thickhline
\multicolumn{1}{c|}{Metric} & Definition\\
\hline
    Abs-rel $\downarrow$ & $\frac{1}{n}\sum|d - d_{gt}|/d_{gt}$\\
    Abs-diff $\downarrow$& $\frac{1}{n}\sum|d - d_{gt}|$ \\
    Abs-inv $\downarrow$ & $\frac{1}{n}\sum|\frac{1}{d} - \frac{1}{d_{gt}}|$ \\
    Sq-rel$\downarrow$& $\frac{1}{n}\sum|d - d_{gt}|^2/d_{gt}$ \\
    RMSE $\downarrow$& $\sqrt{\frac{1}{n}\sum|d - d_{gt}|^2}$ \\
    $\delta < 1.25$ $\uparrow$&  $\frac{1}{n}\sum(\max(\frac{d}{d_{gt}}, \frac{d_{gt}}{d}) < 1.25)$\\
\hline
\end{tabular}
\end{center}
\vspace{-0.2cm}
\caption{\textbf{Depth-prediction metrics.} $n$ is the number of depth pixels, $d$ and $d_{gt}$ are the predicted and ground truth depth values.}
\label{tab:metrics}
\end{small}
\vspace{-0.3cm}
\end{table}

{
    \small
    \bibliographystyle{ieeenat_fullname}
    \bibliography{main}
}